\pdfoutput=1

\documentclass[11pt]{article}

\usepackage[final]{acl}

\usepackage{times}
\usepackage{latexsym}
\usepackage{hyperref}
\usepackage{color, colortbl}
\definecolor{Gray}{gray}{0.9}
\usepackage{array}
\usepackage{booktabs}
\usepackage{multirow}
\usepackage{amsmath}
\usepackage{amssymb}
\usepackage{graphicx}
\usepackage{enumerate}
\usepackage{diagbox}
\usepackage{mathtools}
\usepackage{paralist}
\usepackage{tabularx}
\usepackage{lipsum}
\usepackage{ragged2e}
\usepackage{url}
\usepackage{enumitem}
\usepackage{courier}
\graphicspath{{images/}}
\usepackage[T1]{fontenc}

\usepackage[utf8]{inputenc}

\usepackage{microtype}

\usepackage{inconsolata}
\newcommand*{\affaddr}[1]{#1} 
\newcommand*{\affmark}[1][*]
{\textsuperscript{#1}}

\newcommand{\name}[1]{\textsc{IntentionQA}}
\newcommand{\taskone}[1]{\textsc{IntentUnderstand}}
\newcommand{\tasktwo}[1]{\textsc{IntentUtilize}}

\newcommand{\secvsabove}{\vspace{-0mm}}
\newcommand{\secvsbelow}{\vspace{-0mm}}
\newcommand{\subsecvs}{\vspace{-0mm}}

\newcommand{\paravs}{\vspace{-2mm}}

\newcommand{\itemvs}{\vspace{-0.1in}}

\title{\name{}: A Benchmark for Evaluating Purchase Intention Comprehension Abilities of Language Models in E-commerce}

\author{
Wenxuan Ding\affmark[1]\thanks{\quad Equal Contribution} ,
Weiqi Wang\affmark[1]$^{*}$, 
Sze Heng Douglas Kwok\affmark[1],
Minghao Liu\affmark[1],\\
\textbf{
Tianqing Fang\affmark[1],
Jiaxin Bai\affmark[1],
Xin Liu\affmark[2],
Changlong Yu\affmark[2],
Zheng Li\affmark[2]},\\
\textbf{
Chen Luo\affmark[2],
Qingyu Yin\affmark[2],
Bing Yin\affmark[2],
Junxian He\affmark[1],
Yangqiu Song\affmark[1]}\\
\affaddr{\affmark[1]Department of Computer Science and Engineering, HKUST, Hong Kong SAR, China}\\
\affaddr{\affmark[2]Amazon.com Inc, Palo Alto, CA, USA} \\
\texttt{wdingaj@connect.ust.hk, \{wwangbw, junxianh, yqsong\}@cse.ust.hk}\\
\texttt{\{xliucr, changlyu, amzzhe, cheluo, qingyy, alexbyin\}@amazon.com}
}

\begin{document}
\maketitle
\begin{abstract}
Enhancing  Language Models' (LMs) ability to understand purchase intentions in E-commerce scenarios is crucial for their effective assistance in various downstream tasks. However, previous approaches that distill intentions from LMs often fail to generate meaningful and human-centric intentions applicable in real-world E-commerce contexts.  
This raises concerns about the true comprehension and utilization of purchase intentions by LMs.
In this paper, we present \name{}, a double-task multiple-choice question answering benchmark to evaluate LMs' comprehension of purchase intentions in E-commerce.
Specifically, LMs are tasked to infer intentions based on purchased products and utilize them to predict additional purchases. 
\name{} consists of 4,360 carefully curated problems across three difficulty levels, constructed using an automated pipeline to ensure scalability on large E-commerce services. Human evaluations demonstrate the high quality and low false-negative rate of our benchmark.
Extensive experiments across 19 language models show that they still struggle with certain scenarios, such as understanding products and intentions accurately, jointly reasoning with products and intentions, and more, in which they fall far behind human performances.
Our code and data are publicly available at \href{https://github.com/HKUST-KnowComp/IntentionQA}{https://github.com/HKUST-KnowComp/IntentionQA}.


\end{abstract}

\secvsabove
\section{Introduction}
\secvsbelow
Purchase intentions are mental states where agents or humans commit themselves to purchasing the products~\cite{yu2023folkscope}.
Understanding customers' purchase intentions and making reasonable inferences accordingly are crucial for revolutionizing E-commerce services, whose benefits have been demonstrated in myriads of downstream tasks, such as product recommendation~\cite{DBLP:conf/kdd/GrbovicRDBSBS15,DBLP:conf/kdd/ZhaoGHJWL14,DBLP:conf/cikm/LiZC20} and query answering~\cite{DBLP:conf/cikm/ZhaoCY19,DBLP:conf/sigir/HirschGNDK20}. 
However, intention comprehension~\cite{fogassi2005parietal} is a non-trivial task as it involves reasoning with implicit mental states, which are not typically expressed in text or conversations. 
Thus, in the context of E-commerce, extracting purchase intentions from behaviors without explicit external cues has been challenging~\cite{DBLP:conf/icdm/YangT15}.

Recently,~\citet{yu2023folkscope,yu2024cosmo,MIKO} proposed to distill purchase intentions from LLMs, such as OPT~\cite{DBLP:journals/corr/abs-2205-01068}, by leveraging their inherent advantages of generative and commonsense reasoning abilities, as well as being pre-trained on vast textual data including E-commerce knowledge. 
However, recent analyses by~\citet{zhou2024usage} show that LMs struggle to generate meaningful and user-centric intentions. 
Instead, they are biased by over-focusing on similarities among different products' metadata, such as their properties, and often end up regurgitating information from the provided prompts without truly comprehending the underlying purchase intentions.

\begin{figure}[t]
    \centering
    \includegraphics[width=1\linewidth]{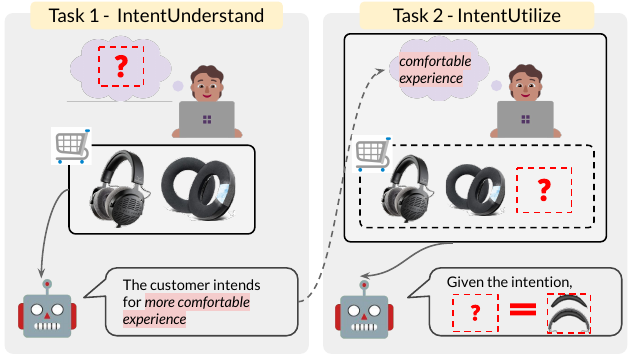}
    \caption{Examples of two tasks in~\name{}.
    Task 1 requires the language model to determine the customer's intention in purchasing two products, and Task 2 involves recommending a product that fulfills the customer's intention and matches their currently purchased product.}
    \label{fig:intro}
\end{figure}

Thus, an important yet under-explored question arises: \emph{Can LMs comprehend the customers' purchase intention and how effective are they in performing such tasks?}
To dive into this, we first break down the comprehension of intention into two key aspects, as shown in Figure~\ref{fig:intro}.
First, we have \textbf{intention understanding}, which evaluates LMs' capacity to accurately infer customers' purchase intentions based on the products bought. 
Second, we consider \textbf{intention utilization}, which investigates LMs' ability to predict additional purchases based on customer's intentions. 
Together, they make up the entire process of intention comprehension and play a significant role in enhancing E-commerce search services.

Although LMs have been extensively used in intention knowledge distillation, their actual performances in this area have not been adequately benchmarked. 
This is because current methods that leverage LMs have been adopting an open-ended generation fashion, which is difficult to consistently evaluate~\cite{DBLP:conf/aaai/GuWY21}.
Additionally, the extensive and constantly growing number of products on E-commerce stores makes it infeasible and expensive to construct human-curated benchmarks.

To address these challenges and benchmark LMs on purchase intention comprehension in E-commerce, we introduce \name{}, a double-task multiple-choice question answering (MCQA) dataset, featuring intention understanding and intention utilization respectively.
\name{} contains 4,360 problems for two tasks and covers varying difficulty levels, allowing for fine-grained evaluation.
The MCQA setting enables using consistent evaluation metrics to assess the LMs' intention comprehension abilities.

Specifically, we design a pipeline that automatically synthesizes QA pairs by transforming human-annotated intentions from FolkScope~\citep{yu2023folkscope}, each involving a pair of co-buy products and the corresponding intention of purchasing them, into questions by masking out the intention or one of the products. 
To achieve this, we define context-based product similarity and intention similarity metrics.
They are computed over ASER~\citep{zhang2020aser}, a large-scale eventuality knowledge graph, which we leverage as a reference for our automatic distractor sampling strategy. 
For each question, we include 3 negative distractors alongside the gold answer through a strict similarity filtering process. 
We then assign difficulty labels to each QA pair based on the product similarity between the co-buy products in the original intention assertion. 
These steps are done without human supervision, enabling our benchmark construction pipeline to generalize and accommodate larger-scale product databases and practical applications.

We further conduct human evaluations to demonstrate the high quality and low false-negative rate of \name{}, followed by extensive experiments across 19 language models with varying sizes and approaches.  
Results demonstrate that the existing language models still struggle with certain scenarios, such as understanding products and intentions accurately, jointly reasoning with the products and intentions, and more.
In the long run, we hope that our benchmark serves as an important cornerstone toward intention-aware E-commerce services that promote integrating intention reasoning abilities into product recommendations.

\begin{figure*}[t]
    \centering
    \includegraphics[width=1\linewidth]{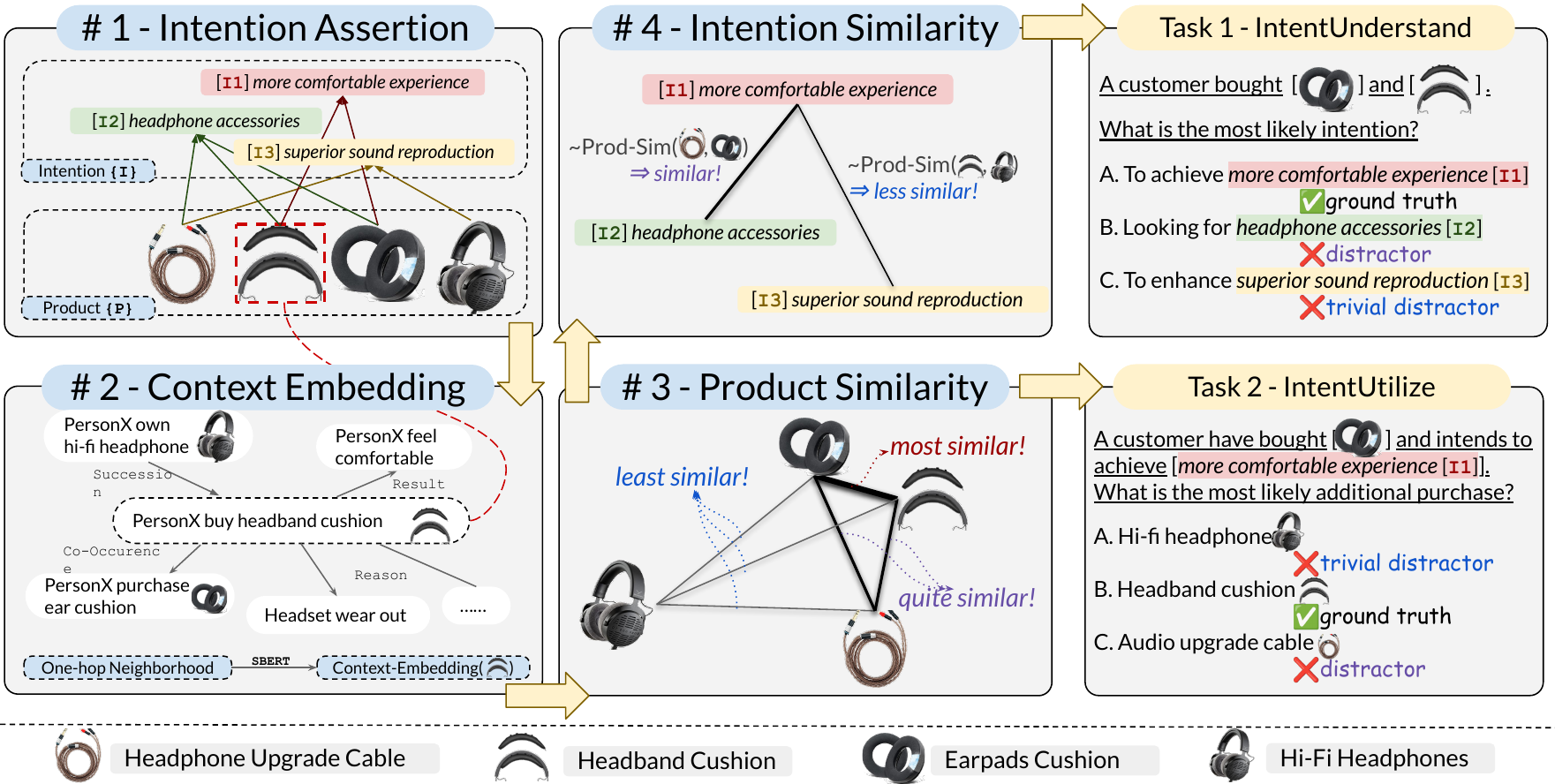}
    \caption{Overview of \name{} and the construction pipeline. We map products from intention assertions to event nodes in ASER (\# 1) and calculate their context embedding with the one-hop neighborhood (\# 2). Product (\#3) and intention (\#4) similarities are then computed accordingly. Products/intentions with higher similarities are represented closer to each other. Negative distractor sampling for Task 1/2 is based on intention/product similarity respectively.
    }
    \label{fig:overview}
\end{figure*}

\secvsabove
\section{Related Works}
\secvsbelow

\subsection{Intention Discovery with Large Language Models in E-commerce}
\subsecvs
Understanding intentions with language models have been studied in various domains, such as smoothing chatbox conversations~\citep{ouyang2022training}, enhancing web search~\citep{zhang2019generic}, and more. 
In the E-commerce domain, understanding customers' purchase intentions benefits various downstream tasks~\citep{koo2010interactional, xu2024MIND, EcomScript}, such as automated on-call customer support \citep{goyal2022end}, recommendation systems \citep{dai2006detecting,qian2023hutcrs,jung2023towards}, product question answering \citep{deng2023product,yu2018product}.
While~\citet{yu2023folkscope,yu2024cosmo} proposed leveraging the generation abilities of LLMs to distill purchase intentions from co-buy records,~\citet{zhou2024usage} showed that LLMs struggle with generating meaningful intentions or understanding user-centric intentions. 
In this work, we construct \name{}, a benchmark to evaluate LMs' intention comprehension abilities by selecting highly typical intentions in previously available resources and provide insights for human-centric intention comprehension.

\subsecvs
\subsection{Benchmarking (Large) Language Models}
\subsecvs
Since the emergence of (L)LMs, various studies have explored their capabilities in various domains, including temporal reasoning~\cite{DBLP:conf/acl/TanNB23}, causal reasoning~\cite{DBLP:conf/eacl/ChanCWJFLS24}, commonsense reasoning~\cite{DBLP:conf/emnlp/JainSA0JD23}, and more~\cite{DBLP:conf/emnlp/QinZ0CYY23,ding2023knowledge,DBLP:conf/nips/BaiLW0S23,feng-etal-2024-dont}. 
These benchmarks have made significant contributions to the understanding of large language models, assessing their performance across different parameters and prompting methods. 
However, in the field of E-commerce, existing works primarily leverage LLMs with explicit instruction-tuning~\cite{DBLP:conf/aaai/LiMWHJ0X0J24}, while neglecting the infeasibility of directly applying LLMs in a generalizable manner. 
Furthermore, current evaluation benchmarks in E-commerce primarily emphasize product and session comprehension~\cite{DBLP:conf/nips/0009MLJ0WHLW0LC23}, which overlooks the important aspect of intention comprehension.
In this paper, we step forward by presenting the first benchmark that evaluates the intention comprehension abilities of (L)LMs.

\secvsabove
\section{\name{}}
\secvsbelow
In this section, we introduce \name{}, a Multiple-Choice Question Answering (MCQA) benchmark consisting of two tasks targeting different aspects of purchase intention comprehension and with progressive difficulties, to evaluate the intention understanding and utilization abilities of LMs thoroughly.
\subsecvs
\subsection{Task Definitions}
\subsecvs
We begin by formally defining two tasks associated with~\name{}. For the tasks presented, we specifically refer to purchase intention as the intention that drives the customer to buy a pair of products together.
\paragraph{Task 1: \taskone{}} The first task examines whether LMs can infer the purchase intentions correctly given a real-world record of the products bought. 
Formally, given a pair of co-buy products $p_1$, $p_2$, LMs are tasked with selecting the most likely purchase intention $i^*$ from a list of candidate options $\mathcal{I} = [i_1, i_2,\ldots, i_{|\mathcal{I}|}]$.
\paragraph{Task 2: \tasktwo{}} The second task looks further into the capacity of LMs to utilize purchase intention for the product recommendation process. 
We approach this by examining their abilities to predict the most likely additional purchase based on customer intention. 
Specifically, given the purchase intention $i^*$ and one product that has been \underline{B}ought $p^{B}$, the LMs are tasked with selecting the most likely 
\underline{A}dditional purchase $p^{A*}$ from a list of candidate options $\mathcal{P^A} = [p^A_1, p^A_2,\ldots, p^A_{|\mathcal{P^A}|}]$.

\subsecvs
\subsection{Source Intention Collection and Context Augmentation}
\subsecvs
We collect co-buy products and intention assertions from FolkScope~\cite{yu2023folkscope} as our source data.
FolkScope is an intention knowledge base that is constructed by distilling knowledge from a pre-trained large language model, OPT~\cite{DBLP:journals/corr/abs-2205-01068}. 
It associates customers' co-purchase behaviors with their purchase intentions, as shown in the upper left part (\# 1) of Figure~\ref{fig:overview}. 
Two scores are also assigned to each intention, indicating its plausibility and typicality.
To accommodate our tasks, we preprocess FolkScope by filtering and retaining plausible assertions with typicality scores above 0.5. 
This is to minimize the number of overly-general intentions, which may be plausible for most products but are not specifically related to the given products.
Including these intentions in~\name{} could lead to many false negative distractors, which harms the quality of our QA pairs.

Since we are aiming for automatic QA pair construction, determining the similarity between different intentions and products can serve as powerful hints in selecting appropriate distractors given a correct answer. 
However, relying solely on product metadata and corresponding purchase behavior falls short of capturing the similarity between intentions, as similar or identical intentions can align with multiple products.
To address this limitation and enhance the sampling of distractors while reducing the occurrence of false-negative distractors, we introduce a method to augment customers' purchase behavior. 
This is achieved by retrieving additional relevant context from ASER~\cite{DBLP:conf/www/ZhangLPSL20,zhang2020aser}, a large-scale eventuality knowledge graph that covers billions of commonly seen eventualities. 
We choose ASER for its extensive knowledge coverage, prompt consistency, and cost-effectiveness compared to API-accessed LLMs (Appendix~\ref{appendix:fs_aser}).

Specifically, we first consider the purchasing event as an eventuality and design heuristic rules to align it with nodes in ASER. 
Formally, we denote ASER as $\mathcal{G} = \{(h,r,t)|h,t \in \mathcal{V}, r \in \mathcal{R}\}$, where $\mathcal{V}$ and $\mathcal{R}$ are the sets of nodes and relations in ASER. In ASER, the nodes are events and the edges are relationships between different events. For example, a node in ASER could be ``\texttt{PersonX feel comfortable}'' and this node is the \emph{result} of the node ``\texttt{PersonX buy headband cushion}''. Tile \# 2 in Figure~\ref{fig:overview} is an example subgraph from ASER.

Inspired by recent works in conceptualization~\cite{DBLP:conf/acl/WangFXBSC23,DBLP:conf/emnlp/WangF0XLSB23,MARS,DBLP:journals/corr/abs-2406-10885,DBLP:journals/ai/HeFWS24}, we simplify the product name $p$ by instructing ChatGPT to conceptualize it into three plausible categories $\mathcal{C}(p) = [c_1, c_2, c_3]$, using prompts presented in Table~\ref{tab:prompt-sim}.
For example, \textit{IPhone 14} can be conceptualized as a \textit{phone}, \textit{communication device}, and \textit{Apple product}. 
This augmentation expands the semantic coverage of the purchasing event, increasing the likelihood of finding relevant nodes in ASER.

Next, we design natural language templates (Appendix~\ref{sec:prompt}) to convert noun phrases of conceptualized product categories into purchasing events $\mathcal{E(C)} = [e_1, e_2, \ldots, e_{|\mathcal{E}|}]$. 
These events are then matched against nodes in ASER to identify overlapping ones through strict string matching. 
Formally, we denote ASER as $\mathcal{G} = \{(h,r,t)|h,t \in \mathcal{V}, r \in \mathcal{R}\}$, where $\mathcal{V}$ and $\mathcal{R}$ are the sets of nodes and relations in ASER. 
We denote the set of matched nodes for $p$ in ASER as $\mathcal{VE} = \mathcal{E(C)} \cap \mathcal{V} = [ve_1, ve_2, \ldots, ve_{|\mathcal{VE}|}]$. 
Next, we compute the sentence embedding of edges in the one-hop neighborhood of each node in $\mathcal{VE}$ using Sentence-BERT~\cite{DBLP:conf/emnlp/ReimersG19}. 
The context embedding $CE(p)$ is then computed by averaging these embeddings, which serves as the semantic representation of relevant contexts for purchasing the product $p$.

\subsecvs
\subsection{Product and Intention Similarity} 
\label{sec:similarity}
\subsecvs
With the context embeddings of all products computed, they contain valuable background knowledge about purchasing events associated with each product. 
This includes edges from ASER that capture cause-effect relationships (``Reason'' and ``Result''), event precedence and succession (``Precedence'' and ``Succession''), and other relevant information. 
Intuitively, similar products should have similar contextual information in ASER, and vice versa, as illustrated in Tile \# 3 of Figure~\ref{fig:overview}. 
Thus, we define the similarity between purchasing events of $p_1$ and $p_2$ as follows:
\begin{align*}
\textit{Sim}^{(p)}(p_1, p_2) = \textit{cos\_sim}(CE(p_1), CE(p_2))
\end{align*}
where \textit{cos\_sim} is the cosine similarity between embeddings from Sentence-BERT.

Similarly, we define the similarity between two intentions ($i_1$, $i_2$) in FolkScope by leveraging their corresponding purchase events associated with ASER as follows:
\begin{align*}
\textit{Sim}^{(i)}(i_1, i_2) = \min_{m= 1, 2; n = 1,2}\{\textit{Sim}^{(p)}(p^{(i_1)}_m, p^{(i_2)}_n)\}
\end{align*}
where $p^{(i)}_m$ is the $\text{m}_\text{th}$ product linked to intention $i$ (Tile \# 4 of Figure~\ref{fig:overview}).

\subsecvs
\subsection{Distractor Sampling and QA Construction}
\subsecvs
Finally, we design~\name{} as a multiple-choice QA benchmark and design specific rules to transform intention assertions in FolkScope into question and gold answer pairs. 
Each gold answer is accompanied by three distractors, i.e., $|\mathcal{I}|,|\mathcal{P^A}|=4$. 
For each task, we propose its unique distractor sampling strategy specifically designed for the task objective, based on the similarity scores defined in \S \ref{sec:similarity}.

\paragraph{Task 1: \taskone{}}
\taskone{} targets LMs' ability to accurately infer purchase intentions based on the products bought by a customer.
We convert the intention assertions from FolkScope to questions by masking out the intentions. 
These masked intentions are then treated as gold answers, denoted as $i^*$.
To obtain the distractor intentions $\mathcal{I^-} = [i_1^-, i_2^-,i_3^-]$, we randomly select intentions from FolkScope whose intention-similarity score with $i^*$ ($\textit{Sim}^{(i)}(i^*, i^-)$) fall within $[0.6,0.9]$.
The lower bound of the range filters out trivial distractors, while the upper bound minimizes the false negative rate in the resulting benchmark. 

\paragraph{Task 2: \tasktwo{}}
\tasktwo{} evaluates the LMs' ability to utilize intentions to predict future purchase behavior. 
Specifically, we formulate the task as providing LMs with one product that the customer has bought and the corresponding intention, and task LMs with predicting the most likely purchase accompanied by the purchased product.
Questions for \tasktwo{} are obtained by masking out one of the products ($p^{A*}$) in each intention assertion of FolkScope. 
The distractor products ($p^-_i, i = 1,2,3$) are randomly selected from those products whose product-similarity score $\textit{Sim}^p(p^{A*}, p^{A-})$ falls within $[0.7,0.9]$. 
Threshold values for both tasks are determined through observations of the distribution and preliminary experiments.

\paragraph{Difficulty Labeling}
To allow for fine-grained evaluation, we categorize each question into three difficulty levels. 
Intuitively, intention assertions with high product-similarity scores among co-buy products result in relatively easy problems.
This is based on the assumption that understanding just one product is sufficient for comprehending the corresponding intention, without necessitating reasoning about the relationship between the products.
Conversely, intention assertions with low product-similarity scores contribute to harder problems as they require comprehending both products and their corresponding intentions, as well as reasoning about the potentially complementary relationship between the products.

Therefore, we categorize the problems based on the product-similarity scores of co-buy products in the original intention assertion. 
Specifically, problems with a product-similarity score within the range of $[0.85,1]$ are classified as easy problems, those within the range of $[0.6,0.85)$ are considered medium, and those within the range of $[0,0.6)$ are classified as hard problems.
These thresholds are determined based on distributions and human observations of problem difficulty.

\paragraph{Quality Control}
After a preliminary human evaluation of the resulting QA pairs for both tasks, we observe that for the hard subset problems in \textsc{Task 1} the correctness rate is relatively low and the false-negative rate is relatively high. Therefore, we manually review every problem in this subset and discard those with incorrect gold answers or false-negative options. (Details in Appendix~\ref{appendix:quality_control})
\secvsabove
\section{Benchmark Evaluations}
\secvsbelow
\subsection{Statistics}
We initially construct~\name{} by using 2,315 intention assertions sourced from FolkScope. 
They are selected by filtering those with high plausibility and typicality scores and whose both products can be aligned with purchasing event nodes of ASER. 
We then construct 4,360 problems for both tasks in \name{}, with each problem labeled with difficulty accordingly. 
The benchmark statistics are reported in Table \ref{tab:benchmark_statistics}.

\begin{table}[t]
\small
\centering
\begin{tabular}{@{}lcccc@{}}
\toprule[1.5pt]
       \multirow{3}{*}{Subset}& \multicolumn{2}{c}{\textsc{Task 1}}& \multicolumn{2}{c}{\textsc{Task 2}}\\ 
       \cmidrule(lr){2-3} \cmidrule(lr){4-5}
       & \multicolumn{1}{c}{\#Q} & \multicolumn{1}{c}{Avg. $Sim^p$} & \multicolumn{1}{c}{\#Q} & \multicolumn{1}{c}{Avg. $Sim^p$} \\
       \midrule[0.75pt]
easy   &1703&0.972&1625& 0.971\\
medium &424 &0.740&385 & 0.744\\
hard   &90 &0.530&133 & 0.514\\
\midrule[0.75pt]
Average  &2217&0.905&2143&0.902\\ 
\bottomrule[1.5pt]
\end{tabular}%
\caption{Statistics of the \name{}. We report the number of questions (\#Q) and the average product-similarity scores between the co-buy products among all intentions (Avg. $Sim^p$) within each difficulty subset.}
\label{tab:benchmark_statistics}
\end{table}

\subsection{Human Evaluations}
To evaluate the effectiveness of our benchmark construction pipeline and assess the quality of our constructed QA benchmark, we conduct human annotation to evaluate (1) the correctness of product conceptualization by ChatGPT and (2) the quality of the QA pairs in both tasks.
\subsubsection{Annotation Details}
We recruit human annotators from the Amazon Mechanical Turk service for human evaluation.

For product conceptualization, 
results show that \textbf{$89.4\%$} of products are reasonably conceptualized, demonstrating the strong product understanding ability of ChatGPT and validating the feasibility of leveraging its generative power to aid our benchmark construction process.

For the quality of resulting QA pairs, we randomly sample 300 QA pairs and ask the annotators to assess the quality of these problems, including the correctness of ground truth options (\emph{Correct}) and assess the false-negativeness of the distractor options by determining whether a distractor option is superior to or equally plausible as the ground truth option (\emph{F-Neg}). 

\begin{table}[t]
\small
\centering
\begin{tabular}{@{}lcccc@{}}
\toprule[1.5pt]
       \multirow{3}{*}{Subset}& \multicolumn{2}{c}{\textsc{Task 1}}& \multicolumn{2}{c}{\textsc{Task 2}}\\ 
       \cmidrule(lr){2-3} \cmidrule(lr){4-5}
       & \multicolumn{1}{c}{Correct} & \multicolumn{1}{c}{F-Neg} & \multicolumn{1}{c}{Correct} & \multicolumn{1}{c}{F-Neg} \\
       \midrule[0.75pt]
easy   &96.07&2.77&98.20&1.20\\
medium &94.00&2.67&92.59&4.32\\
hard   &100.00&0.00&100.00&0.00\\
\midrule[0.75pt]
Average  &96.00&2.56&97.33&1.67\\
\bottomrule[1.5pt]
\end{tabular}%
\caption{Annotated correctness (Correct; \%) and false-negative rate (F-Neg; \%) of 600 randomly sampled QA pairs from two tasks.}
\label{tab:expert_annotation}
\end{table}

\subsubsection{Annotation Results}
We report the annotation results in Table~\ref{tab:expert_annotation}. 
We find that~\name{} exhibits high correctness rates among ground truth options.
Meanwhile, the low false-negative rates demonstrate the high quality of both tasks.
Both statistics validate the reliability of our automatic QA construction pipeline and the quality of the resulting~\name{} benchmark.

\secvsabove
\section{Experiments and Analysis}
\secvsbelow
\begin{table*}[t]
    \small
    \setlength{\tabcolsep}{4.9pt}
    \resizebox{1\linewidth}{!}{
	\centering
	\begin{tabular}{llcccccccc}
	\toprule[1.5pt]
        \multirow{2}{*}{\textbf{Methods}}&\multirow{2}{*}{\textbf{Backbone}}&\multicolumn{4}{c}{\textbf{\taskone{}}} &\multicolumn{4}{c}{\textbf{\tasktwo{}}}\\
        \cmidrule(lr){3-6}\cmidrule(lr){7-10}
	&&\textbf{Easy}&\textbf{Medium} &\textbf{Hard} &\textbf{Avg.}&\textbf{Easy}&\textbf{Medium} &\textbf{Hard} &\textbf{Avg.} \\
            \midrule[0.75pt]
            \textbf{Random} &  - & 25.00 & 25.00 & 25.00 & 25.00 & 25.00 & 25.00 & 25.00 & 25.00 \\
            \textbf{Majority Vote} & - & 26.37	& 25.24	& 26.27	& 25.00	& 26.09	& 28.57 & 28.57 & 26.60 \\
            \midrule[0.75pt]
            \multirow{5}{*}{\textbf{PTLM}}&RoBERTa-Large \scriptsize{\textit{214M}} &41.46&41.98&38.98&41.43&54.95&35.06&30.08&49.84\\
            & DeBERTa-v3-Large \scriptsize{\textit{435M}}&36.40&38.72&37.62&36.90&26.52&29.35&32.33&27.39\\
            &T5-v1.1-xxl \scriptsize{\textit{11B}}&24.84&25.47&23.91&24.92&26.71&26.23&25.56&26.55\\
            &Flan-T5-xxl \scriptsize{\textit{11B}}&\underline{75.98}&\underline{73.58}&75.00&75.48&\underline{79.26}&\underline{81.82}&\underline{81.95}&\underline{79.89}\\
            &T0-pp \scriptsize{\textit{11B}}&71.70&68.87&69.57&71.07&77.11&76.10&78.20&76.99\\
		  \midrule[0.75pt]
            \multirow{5}{*}{\textbf{Commonsense}}&HyKAS \scriptsize{\textit{435M}}&71.81&67.17&46.69&69.61&47.02&45.97&48.12&46.90\\
            &CAR \scriptsize{\textit{435M}}&73.69&71.46&54.38&72.20&36.18&43.12&44.36&37.94\\
            &CANDLE \scriptsize{\textit{435M}}&\underline{74.34}&70.75&52.54&72.52&35.94&43.90&43.61&37.84\\
            &VERA \scriptsize{\textit{11B}}&69.82&70.52&61.02&69.49&59.20&58.18&64.66&59.36\\
            &VERA-CANDLE \scriptsize{\textit{11B}}&70.59&\underline{71.33}&\underline{63.41}&\underline{70.02}&\underline{62.18}&\underline{60.13}&\underline{66.13}&\underline{61.81}\\
            \midrule[0.75pt]
            \multirow{15}{*}{\textbf{Open LLM}}&LLaMA2-7B&22.47&26.24&22.37&23.18&26.42&27.87&29.03&26.84\\
            &LLaMA2-7B-chat&64.98&66.54&57.38&64.93&\underline{59.90}&54.86&47.37&58.04\\
            &LLaMA2-13B&24.21&27.70&25.00&24.91&27.92&30.59&28.03&28.40\\
            &LLaMA2-13B-chat&69.63&63.96&62.50&68.21&45.53&41.95&39.71&44.52\\
            &Gemma-2B&21.73&23.87&19.51&22.08&30.66&30.63&30.99&30.67\\
            &Gemma-2B-instruct&48.77&47.23&53.41&48.67&39.45&39.15&38.17&39.32\\
            &Gemma-7B&50.94&50.86&47.19&50.77&26.75&30.19&31.20&27.65\\
            &Gemma-7B-instruct&65.55&64.31&61.04&65.13&33.18&36.01&41.51&34.20\\
            &Mistral-7B-instruct-v0.1&53.49&55.04&53.64&53.80&26.18&28.27&28.57&26.70\\
            &Mistral-7B-instruct-v0.2&\underline{76.57}&\underline{74.53}&\underline{72.83}&\underline{76.03}&59.78&\underline{62.60}&\underline{65.41}&\underline{60.64}\\
            &Falcon-7B&24.19&20.52&25.00&23.52&25.40&25.45&27.82&25.56\\
            &Falcon-7B-instruct&24.54&22.17&28.26&24.25&26.15&28.05&26.32&26.50\\
            &Vicuna-7B-v1.5&57.13&57.08&55.43&57.05&27.88&30.13&23.31&28.00\\
            &Phi-2 \scriptsize{\textit{3B}}&33.24&37.97&33.70&34.16&26.71&28.57&28.57&27.16\\
            &Alpaca-LLaMA-7B&48.97&46.93&44.57&48.40&50.15&46.49&37.59&48.72\\
            \midrule[0.75pt]
            \multirow{2}{*}{\textbf{Open LLM + MIND}}&LLaMA2-7B-chat&65.78&64.61&55.75&66.15&59.43&57.13&60.03&59.04\\
            &Mistral-7B-instruct-v0.2&\underline{78.57}&\underline{74.31}&\underline{80.89}&\underline{76.97}&\underline{61.14}&\underline{65.42}&\underline{62.16}&\underline{62.02}\\
            \midrule[0.75pt]
            \multirow{6}{*}{\textbf{LLM API}}
            &ChatGPT 
            &75.06&73.76&77.17&74.90&80.74&76.62&68.42&79.23\\
            &ChatGPT (CoT) &76.07&74.53&72.83&75.64&78.89&75.32&78.20&78.21\\
            &ChatGPT (CoT-SC) &76.51&73.82&71.74&75.80&85.72&77.14&82.71&83.99\\
            &GPT-4 &78.12&\underline{\textbf{75.41}}&73.91&77.43&\underline{\textbf{86.03}}&\underline{\textbf{82.34}}&\underline{\textbf{84.96}}&\underline{\textbf{85.30}}\\
            &GPT-4 (CoT) &77.43&73.11&\underline{\textbf{80.43}}&76.73&83.57&79.74&82.71&82.83\\
            &GPT-4 (CoT-SC) &\underline{\textbf{78.80}}&72.88&\underline{75.00}&\underline{\textbf{77.51}}&84.00&80.78&\underline{\textbf{84.96}}&83.48\\
            \midrule[0.75pt]
            \midrule[0.75pt]
            \textbf{Human} & - &89.96&90.00&100.00&90.67&95.50&85.19&100.0&94.00\\
		\bottomrule[1.5pt]
	\end{tabular}
 }
	\caption{Evaluation results (Accuracy\%) of various language models on both tasks of the \name{} benchmark.
 The best performances within each category are \underline{underlined} and the best among all baselines are \textbf{bold-faced}.}
    \label{tab:main_eval_results}
\end{table*}

\subsecvs
\subsection{Baseline Selection and Setup}
\subsecvs
\paragraph{Evaluation Metric} We use accuracy as the evaluation metric, which is quantified by the percentage of QA pairs that a language model answers correctly in~\name{}.\paravs
\paragraph{Model Selection}
We evaluate a wide range of (L)LMs in four categories: 
\textbf{(1) \textsc{PTLM}:} We evaluate several pre-trained language models, including RoBERTa~\cite{DBLP:journals/corr/abs-1907-11692}, DeBERTa-v3~\cite{DBLP:conf/iclr/HeGC23}, T0~\cite{DBLP:conf/iclr/SanhWRBSACSRDBX22}, T5~\cite{DBLP:journals/jmlr/RaffelSRLNMZLL20}, and Flan-T5~\cite{DBLP:journals/corr/abs-2210-11416}. 
\textbf{(2) \textsc{Commonsense}:} We also evaluate PTLMs with commonsense knowledge injected, including HyKAS~\cite{DBLP:conf/aaai/MaIFBNO21}, CAR~\cite{DBLP:conf/emnlp/WangF0XLSB23}, VERA~\cite{DBLP:conf/emnlp/0010WWS0H23}, CANDLE~\cite{DBLP:journals/corr/abs-2401-07286}, and VERA-CANDLE~\cite{DBLP:journals/corr/abs-2401-07286}. 
\textbf{(3) \textsc{Open LLM}:} We then evaluate representative open-sourced LLMs of varying sizes and versions in zero-shot settings as well as after fine-tuning on intention knowledge (\textsc{Open LLM + Mind}, details in \S\ref{sec:mind}). 
These models cover LLaMA2~\cite{DBLP:journals/corr/abs-2307-09288}, Gemma~\cite{DBLP:journals/corr/abs-2403-08295}, Mistral~\cite{DBLP:journals/corr/abs-2310-06825}, Falcon~\cite{DBLP:journals/corr/abs-2311-16867}, Vicuna~\cite{DBLP:conf/nips/ZhengC00WZL0LXZ23}, Phi-2~\cite{DBLP:journals/corr/abs-2306-11644}, and Alpaca~\cite{alpaca,DBLP:conf/acl/WangKMLSKH23}.
\textbf{(4) \textsc{LLM API}:} Finally, we adopt Chain-of-Thought prompting (\textsc{CoT};~\citealp{wei2022chain}) and CoT with Self-Consistency (\textsc{CoT-SC};~\citealp{wang2022self}) together with zero-shot prompting to assess ChatGPT~\cite{openai2022chatgpt} and GPT-4~\cite{GPT4}. 
The sampling temperature $\tau$ is set to $0.1$ by default.
$5$ \textsc{CoT} responses are sampled with $\tau$ set to $0.7$ under \textsc{CoT-SC}.
\textsc{Random} and \textsc{Majority} voting are also added as baselines to demonstrate the characteristic of~\name{}. 
\textsc{Human} performance is calculated based on annotation results of 600 randomly selected QA pairs from both tasks.
(See prompts in Table~\ref{tab:prompt-task1} and~\ref{tab:prompt-task2})
\paravs
\subsecvs
\subsection{Results}
\subsecvs
The results of all models are presented in Table \ref{tab:main_eval_results}.
From the results, we observe that:

\noindent\textbf{Commonsense knowledge does help in intention comprehension.} 
Models injected with commonsense knowledge showcase comparable performance to significantly larger models. 
Specifically, CAR and CANDLE (435M) achieve $96.64\%$ of the performance of Flan-T5-xxl (11B) in \taskone{}, despite being 25 times smaller.
This demonstrates the effectiveness of incorporating commonsense knowledge in improving intention comprehension in the E-commerce domain.

\noindent\textbf{\tasktwo{} is more challenging.} 
For approximately all models, excluding ChatGPT and GPT-4, that exhibit above \textsc{Random} performances in \taskone{}, their performances drop significantly when evaluated on \tasktwo{}, with an average accuracy gap of $14.20\%$.
While \taskone{} involves understanding the purchase intention behind a single pair of products, \tasktwo{} requires product understanding of all candidate options as well as reasoning with potential intentions behind four pairs of products. 
This expanded reasoning scope and higher demand for product understanding pose challenges for these models, as their training data may be limited in terms of the variety and quantity of products included.
However, ChatGPT and GPT-4 excelled in both tasks, presumably due to their stronger product reasoning abilities.

\noindent\textbf{Intention comprehension abilities of current models are still far from perfect.} 
Although various models perform considerably better than \textsc{Random} guessing, there remains a substantial gap between their performance and that of humans.

\subsecvs
\subsection{Performances Across Intention Types}
\begin{figure}[t]
    \centering
    \includegraphics[width=1\linewidth]{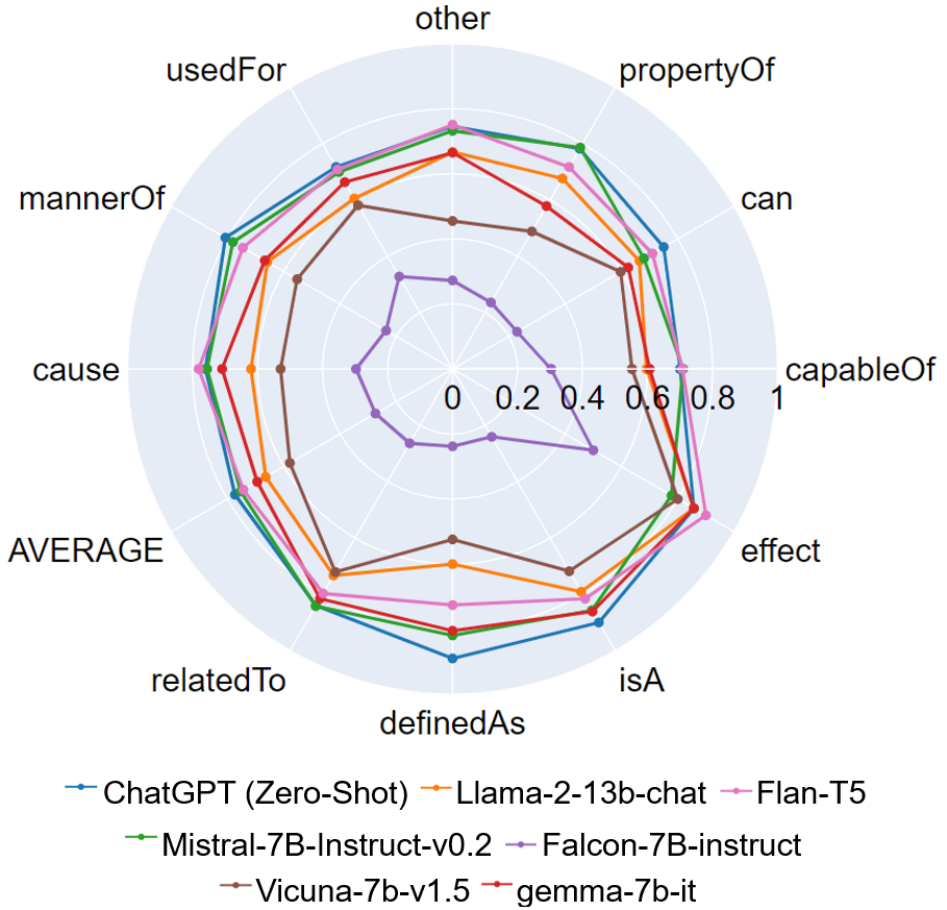}
    \caption{Performances of various language models in comprehending intentions with different relations.}
    \label{fig:byrelation}
\end{figure}

\subsecvs
To further investigate the reasons why language models fail in intention comprehension, we conduct a more fine-grained analysis by delving into intentions with different commonsense relations grounded in ConceptNet~\cite{DBLP:conf/aaai/SpeerCH17}. 
Specifically, we construct a sibling QA set using our proposed pipeline, with the only additional constraint being that the distractor options share the same relation type as the ground truth option. 
From the results presented in Figure \ref{fig:byrelation}, all the evaluated language models are more effective in understanding the product definition, with an average of $70.47\%$ across relations \texttt{isA}, \texttt{definedAs}, and \texttt{relatedTo}. 
However, a performance decline of $6.69\%$ is observed in relations that require a deeper understanding of the cause and effect behind the purchasing event, such as \texttt{capableOf} and \texttt{cause}.

\secvsabove
\subsection{Error Analysis}
\secvsbelow
In this section, we randomly sample 120 questions that GPT-4 answers incorrectly from \name{} and categorize the errors by asking experts to annotate them manually. (Details in Appendix~\ref{appendix:err_analysis})

Among 60 annotated error samples from \taskone{}, we found:
\begin{itemize}[leftmargin=*]
    \item $56.7\%$ errors are caused by failing to identify the most typical intention, e.g., choosing ``because the product is of good quality'' instead of ``because the person wants to build a water cooling system.''
    \itemvs
    \item $18.3\%$ errors are due to overarching inference. The selected options, while seemingly plausible, cannot be deduced from the products provided.
    \itemvs
    \item $11.7\%$ errors are due to selecting implausible options. 
    The model selects an option that is irrelevant to the given products or implausible. 
    Cases where the rationales in \textsc{CoT} responses are irrelevant to selected options are also observed.
    \itemvs
    \item $13.3\%$ errors are due to inaccurate understanding of the given products.
    \itemvs

\end{itemize}

\begin{figure}[t]
    \centering
    \includegraphics[width=1\linewidth]{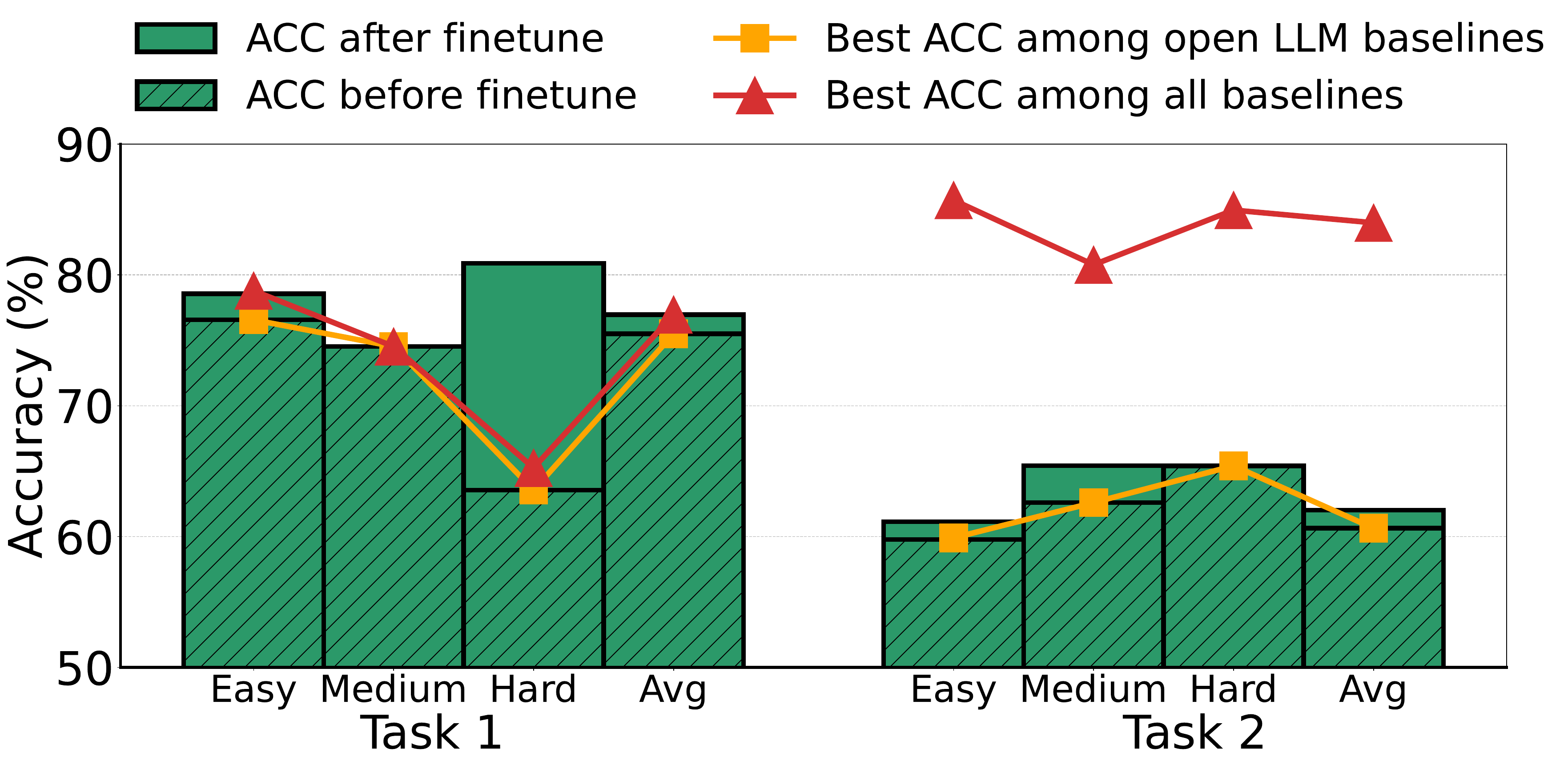}
    \caption{Comparisons between models fine-tuned on intentions from MIND and baseline models achieving top performances.}
    \label{fig:finetune}
\end{figure}

Among 60 annotated error samples from \tasktwo{}, we found:
\begin{itemize}[leftmargin=*]
    \item $40\%$ errors are due to inaccurate understanding of the given intention. For example, the model chooses ``iPod'' under the intention ``because the customer wanted to use them \emph{with} his/her iPod''.
    \itemvs
    \item $38.3\%$ errors are due to inaccurate understanding of the given products. The reasoning in their response demonstrates inaccurate understanding of the purchased products or those in the options. Or, when the intention is not typical enough to filter out distractors, they fail to rely more on the purchased product to select the best option.
    \itemvs
    \item $21.7\%$ errors are due to false-negative distractors or incorrect ground truth answers. 
    \itemvs
\end{itemize}
\subsecvs
\subsection{Transferring from Other Sources}
\label{sec:mind}
\subsecvs
In this section, we explore whether transferring intention knowledge from other sources can aid the model's performance via fine-tuning. 
Specifically, we use MIND, a knowledge base constructed by~\citet{xu2024MIND}, besides FolkScope, as a rich source of purchase intentions. 
MIND is a multi-modal intention knowledge base distilled from LLaVA~\cite{DBLP:conf/nips/LiuLWL23a}, which includes product images in the knowledge generation process. 
To ensure the quality of generated intentions, a human-centric intention filtering module is developed to eliminate implausible and atypical intentions.

To incorporate MIND's intention knowledge, we convert 4,059 sets of co-buy records and their corresponding intentions into an instruction-tuning format. 
We then fine-tune the LLaMA2-7B-chat and Mistral-7B-instruct-v0.2 models on this data using LoRA~\cite{DBLP:conf/iclr/HuSWALWWC22}. 
The results, reported in \textsc{Open LLM + Mind} of Table~\ref{tab:main_eval_results}, reveal an average performance gain of 1.51\% and 1.19\% for two tasks respectively.

Next, we compare the performance of the fine-tuned Mistral-7B-instruct-v0.2 model with the highest accuracy achieved by all \textsc{Open LLM}s and all baselines. 
The trends are shown in Figure~\ref{fig:finetune}. 
Notably, fine-tuning enables Mistral-7B-instruct-v0.2 to achieve performance comparable to that of GPT-4 in~\taskone{}. 
However,~\tasktwo{} remains a challenging task even after fine-tuning. 
This disparity suggests that fine-tuning with intention knowledge facilitates the acquisition of intention understanding abilities, while improving~\tasktwo{} performance requires more effort. 
One possible reason is that~\tasktwo{} places a higher demand on product understanding and reasoning abilities compared to~\taskone{}, which cannot be easily improved by simple knowledge injection.

\secvsabove
\section{Conclusions}
\secvsbelow
In conclusion, this paper presents~\name{}, a double-task MCQA dataset designed to assess the intention comprehension capabilities of LMs. 
Extensive experiments and analyses demonstrate that LMs face significant challenges in certain scenarios, trailing far behind human performance levels, while fine-tuning on external resources brings considerable performance gains. 
We hope our work sheds light on the limitations of current LMs in E-commerce intention understanding tasks and opens up a new paradigm of leveraging LM in E-commerce services.

\section*{Limitations}
We base the negative distractor sampling on similarity filtering with manually selected thresholds. 
While these thresholds are decided after multiple rounds of parameter searches and observation of the resulting data quality and have been validated by the human annotation we conduct, automated threshold tuning methods~\cite{xu2021dash} could be implemented to facilitate this process.

As we build the dataset based on FolkScope, the quality of the latter is upper-bounded by the former. 
Nevertheless, the construction pipeline introduced in this work can be generalized to expand the dataset by incorporating other intention knowledge bases.
Meanwhile, more advanced LLMs have the potential of curating intention knowledge bases with high quality, further boosting the quality of our QA benchmark. 

Since (L)LMs demonstrate strong generative capabilities and commonsense reasoning, it is potentially feasible to leverage models such as ChatGPT to generate contextual information for purchase events. 
However, we rely on the eventuality knowledge graph, ASER, to facilitate the calculation of context embeddings. 
This offers advantages in terms of cost control and the potential to scale up. 
Additionally, the human annotation results of our dataset confirm the effectiveness of leveraging ASER for this purpose.

Our work mainly focuses on intention comprehension in E-commerce, which specifically involves product understanding, purchase intention reasoning, and mental state sharing, as well as commonsense reasoning within the context of intention comprehension in broader domains. While our work studies intention comprehension in a specific domain and holds potential for real-world applications, we believe its findings can offer general insights for broader research.

\section*{Ethics Statement}
\subsection{Offensive Content Elimination}
While we adopt LMs in a generative setting, generating harmful or biased content from them is limited as \name{} is evaluated in multiple-choice question form. 
In most cases, the language models generate a single letter representing the option.
In \textsc{CoT}, the LLMs generate a short rationale and then output the final answer, where the rationale is closely related to the question itself. 
\subsection{Annotations}
The annotators are paid a wage higher than our local law, and the expert annotators are graduate students specializing in natural language processing. 
They have all agreed to participate voluntarily and are well-instructed about the tasks.
\subsection{Licenses}
FolkScope and ASER are released under the MIT license, which grants us access to the datasets for free. Assets including models and tokenizers obtained from Huggingface Hub \footnote{
https://huggingface.co/} are shared via licenses that support research purposes. We will share our code and data under the MIT license, which allows free distribution of our curated assets.
All associated licenses permit user access for research purposes, and we have agreed to follow all
terms of use.

\section*{Acknowledgements}
We thank the anonymous reviewers and the area chair for their constructive comments.
The authors of this paper were supported by the NSFC Fund (U20B2053) from the NSFC of China, the RIF (R6020-19 and R6021-20) and the GRF (16211520 and 16205322) from RGC of Hong Kong. 
We also thank the general support from Amazon.com Inc.

\bibliography{anthology,custom}

\newpage
\appendix

\begin{center}
    {\Large\textbf{Appendices}}
\end{center}
\section{Discussion}
\subsection{Plausibility of leveraging ASER for context augmentation}
\label{appendix:fs_aser}
We decide to use ASER after considering the following factors. First, LLMs accessed via API such as ChatGPT or GPT-4 will incur high costs to generate the context information of each purchase event and therefore inhibit the potential to scale up. Besides, since LLMs are sensitive to prompts, the similarity computed based on their generated context information may be inconsistent. 
Secondly, we observe that open-sourced LMs are relatively weak in understanding the purchase events and generating meaningful context information consistently. Such weakness is further confirmed by their low performance in both tasks. 

However, since ASER is a unified large-scale eventuality knowledge graph, it enjoys a large knowledge coverage and consistency, the similarity computed based on the context information extracted from ASER is fair and consistent. The human annotation also verifies the effectiveness of using ASER.

\subsection{Elaboration on Error Analysis}
\label{appendix:err_analysis}
\paragraph{Overly-general intentions vs. typical intentions}
Overly general intentions are the intentions applicable to almost all products, such as ``because the products are similar to each other'', ``because they are high quality products'', ``because they both are a type of product that he needed''. These intentions are too general and thus lack meaningful specificity.

On the other hand, examples such as ``because they both are capable of producing high quality sound'', ``because they are both used for his laptop'', ``because they both are capable of cooling his CPU'' are more typical intentions and could be more helpful in real-world applications. 

\paragraph{Selection of implausible intentions}
We observe from the errors that the rationale generated with CoT is neither relevant to the options provided nor to the answer the model provides. This may indicate that LLM is distracted by some other information that shares high correlation with the option and thus fails to reason within the provided scenario. We provide an example of such cases in Tabel~\ref{tab:implausibleopt}.

\begin{table*}[t]
\small
\vspace{-0.1in}
\centering
{\def\arraystretch{1.6}
\begin{tabularx}{\textwidth}{>{\raggedright}p{2cm}>{\RaggedRight\arraybackslash}p{12cm}} \toprule[1.5pt]
Task & Prompt \& Response \\ 
\midrule[0.75pt]
Task 1 & 
\underline{[Prompt]}\newline
A customer buys 14 Karat Yellow Gold Round Bead Ball Stud Earrings, 2mm and 14 Karat White Gold Round Bead Ball Stud Earrings, 3mm. What is the most likely intention for buying them?\newline
A. because they both are capable of being worn for a long time.\newline
B. because they both are a type of USB extension cable.\newline
C. because the person wants to use them together.\newline
D. because they both have a property of anti-glare/anti-fingerprint.\newline
Formulate your answer in this way.\newline
Step 1: Give a short and brief rationale by thinking step by step.\newline
Step 2: Answer A or B or C or D only without any other word.\newline
\underline{[Response from GPT-4]}
Step 1: The customer is buying two different types of gold earrings, which indicates they are likely purchasing them for personal use or as a gift. The other options do not relate to earrings or their intended use.\newline
Step 2: C\ \ \ \underline{[Correct answer: A]}\\
\bottomrule[1.5pt]
\end{tabularx}
}
\caption{An example of selection of implausible intention by GPT-4 with CoT prompting. The LLM may be distracted by the property of earrings as being gifts and provide a rationale that is irrelevant to the given options. Although it is hard to use two pairs of earrings together, the LLM fails to select the most likely and plausible intention probably due to its irrelevant rationale.}
\label{tab:implausibleopt}
\end{table*}

\subsection{Quality Control Details}
\label{appendix:quality_control}
The preliminary human annotation over 300 randomly sampled QA pairs shows that the Correctness rate of the \textbf{hard} subset in \textsc{Task 1} is $85.71\%$ and the false-negative rate is $4.76\%$. Therefore, we manually correct this subset by discarding problematic QA pairs. The relatively low correctness rate and high false-negative rate in the hard subset may be because the intention assertions in FolkScope where the two products are less similar suffer from lower quality.

\section{Implementation Details}
\label{appendix:implementation_details}

\subsection{Hyperparameter Settings}
For models in the category of \textsc{LLM API}, we set the \texttt{max\_tokens} to $10$ and $200$ respectively for \textsc{Zero-Shot} and \textsc{CoT} (\textsc{CoT+SC}) prompting. The temperature $\tau$ is set to $0.1$ for non-Self-Consistency decoding and $0.7$ otherwise. 

For models in the category of \textsc{Open LLM}, we use the default setting as presented in Hugging Face.

For fine-tuning LLMs, we use the open-sourced library LLaMA-Factory\footnote{\href{https://github.com/hiyouga/LLaMA-Factory}{https://github.com/hiyouga/LLaMA-Factory}}~\cite{DBLP:journals/corr/abs-2403-13372} to train all models.
All hyperparameters follow the default settings, and a LoRA rank of $\alpha=64$ is used.
We conduct all experiments on a Linux machine with eight NVIDIA V100 GPUs.

\section{Annotation Setup}
We follow annotation setup of~\citet{DBLP:conf/emnlp/FangWCHZSH21,DBLP:conf/www/FangZWSH21,DBLP:conf/emnlp/ShiWFX0LS23} and recruit human annotators from the Amazon Mechanical Turk service for human evaluation.
For strict quality control, we only invite workers satisfying the following requirements: 1) at least 1K HITs approved, and 2) at least 95\% approval rate.
We then host two rounds of qualification rounds using questions sampled from our curated benchmark, with expert-annotated answers.
400 workers are invited in total and around 60 (15\%) of them are selected.

For product conceptualization, we randomly sample 2,000 products and task each annotator to label the plausibility of the generated categories. Specifically, we ask the annotators to assess whether all three generated product categories are reasonable according to the original products.
Each product is annotated by three annotators and the majority vote is taken as the final label.

We then evaluate the resulting QA pairs from \taskone{} and \tasktwo{}, we randomly sample 300 QA pairs and ask the annotators to assess the quality of these problems, including the correctness of ground truth options (\emph{Correct}) and assess the false-negativeness of the distractor options by determining whether a distractor option is superior to or equally plausible as the ground truth option (\emph{F-Neg}). 
Again, we collect three votes for each QA pair and take the majority of them.

\section{Prompts}
\label{sec:prompt}
We report the prompt used for product simplification with ChatGPT in Tabel~\ref{tab:prompt-sim}.

To transfer a conceptualized product into ASER nodes' format, we utilize natural language templates. 
These templates consist of assembling each product category with a subject and a verb that carry semantic meanings related to purchasing. 
Specifically, the subjects we use include: \texttt{PersonX},\texttt{PersonY},\texttt{PeopleX},\texttt{PeopleY};
Similarly, the verbs we employ are: \texttt{buy}, \texttt{shop}, \texttt{purchase}, \texttt{get}, \texttt{obtain}, \texttt{have}, in simple present tense, original form, simple perfect tense, or past tense, with optional articles (\texttt{a}, \texttt{an},\texttt{the},\texttt{1},\texttt{2}) added before the conceptualized product name.
As a result, when a product such as ``iPhone 14'' occurs, we transform it into a list of concise yet semantically complete events that can potentially be matched in ASER. 
For example, one of the transformed events could be ``PersonX bought a phone.''

\begin{table*}[t]
\small
\centering
{\def\arraystretch{1.6}
\begin{tabularx}{\textwidth}{>{\raggedright}p{3cm}>{\RaggedRight\arraybackslash}p{10cm}} \toprule[1.5pt]
Method & Prompt for Product Name Simplification \\ 
\midrule[0.75pt]
\textsc{Zero-Shot} & 
Product name: <\texttt{product}>; What is the category of the product? Generate three possible categories, each in 2 words, separated by a comma.
\\
\bottomrule[1.5pt]
\end{tabularx}
}
\caption{Prompt used to instruct ChatGPT to conceptualize the product name.}
\label{tab:prompt-sim}
\end{table*}

We report the prompts used for \taskone{} and \tasktwo{} in Table~\ref{tab:prompt-task1} and Table~\ref{tab:prompt-task2} respectively.
\begin{table*}[t]
\small
\centering
{\def\arraystretch{1.6}
\begin{tabularx}{\textwidth}{>{\raggedright}p{3cm}>{\RaggedRight\arraybackslash}p{10cm}} \toprule[1.5pt]
Method & Prompt for \taskone{} \\ 
\midrule[0.75pt]
\textsc{Zero-Shot} & 
A customer buys <\texttt{product 1}> and <\texttt{product 2}>. What is the most likely intention for buying them?\newline
A. because <\texttt{intention 1}>\newline
B. because <\texttt{intention 2}>\newline
C. because <\texttt{intention 3}>\newline
D. because <\texttt{intention 4}>\newline
Answer A or B or C or D only without any other word.
\\\midrule[0.75pt]
\textsc{CoT} & 
A customer buys <\texttt{product 1}> and <\texttt{product 2}>. What is the most likely intention for buying them?\newline
A. because <\texttt{intention 1}>\newline
B. because <\texttt{intention 2}>\newline
C. because <\texttt{intention 3}>\newline
D. because <\texttt{intention 4}>\newline
Formulate your answer in this way.\newline
Step 1: Give a short and brief rationale by thinking step by step.\newline
Step 2: Answer A or B or C or D only without any other word.
\\
\bottomrule[1.5pt]
\end{tabularx}
}
\caption{Prompts for \taskone{} with \textsc{Zero-Shot} prompting and \textsc{CoT} respectively.}
\label{tab:prompt-task1}
\end{table*}

\begin{table*}[t]
\small
\centering
{\def\arraystretch{1.6}
\begin{tabularx}{\textwidth}{>{\raggedright}p{3cm}>{\RaggedRight\arraybackslash}p{10cm}} \toprule[1.5pt]
Method & Prompt for \tasktwo{} \\ 
\midrule[0.75pt]
\textsc{Zero-Shot} & 
A customer buys <\texttt{product}>, because <\texttt{intention}>.\newline
What is the customer's most probable additional purchase?\newline
A. <\texttt{product 1}>\newline
B. <\texttt{product 2}>\newline
C. <\texttt{product 3}>\newline
D. <\texttt{product 4}>\newline
Answer A or B or C or D only without any other word.
\\\midrule[0.75pt]
\textsc{CoT} & 
A customer buys <\texttt{product}>, because <\texttt{intention}>.\newline
What is the customer's most probable additional purchase?\newline
A. <\texttt{product 1}>\newline
B. <\texttt{product 2}>\newline
C. <\texttt{product 3}>\newline
D. <\texttt{product 4}>\newline
Formulate your answer in this way.\newline
Step 1: Give a short and brief rationale by thinking step by step.\newline
Step 2: Answer A or B or C or D only without any other word.
\\
\bottomrule[1.5pt]
\end{tabularx}
}
\caption{Prompts for \tasktwo{} with \textsc{Zero-Shot} prompting and \textsc{CoT} respectively.}
\label{tab:prompt-task2}
\end{table*}

\section{Case Study}
\label{appendix:case_study}
We present example questions that GPT-4 successfully answer or fail with \textsc{CoT} for both tasks in Table~\ref{tab:case}.
\begin{table*}[t]
\small
\vspace{-0.1in}
\centering
{\def\arraystretch{1.6}
\begin{tabularx}{\textwidth}{>{\raggedright}p{2cm}>{\RaggedRight\arraybackslash}p{12cm}} \toprule[1.5pt]
Task & Prompt \& Response \\ 
\midrule[0.75pt]
Task 1 & 
\underline{[Prompt]}\newline
A customer buys Belkin USB A/A Extension Cable, USB Type-A Female and USB Type-A Male (6 Feet) and Cable Matters 3-Pack USB Extension Cable (USB Extender Cable/USB 2.0 Extension Cable) - 3 Feet - Available 3FT - 10FT in Length. What is the most likely intention for buying them?\newline
A. because they both are a type of USB extension cable.\newline
B. because they both are a manner of "Audio \& Video Accessories".\newline
C. because they both are related to the topic of "Army Style".\newline
D. because they could both be worn as a bracelet.\newline
Formulate your answer in this way.\newline
Step 1: Give a short and brief rationale by thinking step by step.\newline
Step 2: Answer A or B or C or D only without any other word.\newline
\underline{[Response from GPT-4]}
Step 1: The customer is buying two different USB extension cables, which indicates they need to extend the length of their USB connections. \newline
Step 2: A\ \ \ \underline{[Correct!]}
\\\midrule[0.75pt]
Task 1 & 
\underline{[Prompt]}\newline
A customer buys Barbra Collection Hawaiian Style Kukui Nut Lei Hand Painted Multi-color Turtle 32 Inches and Barbra Collection Hawaiian Style Kukui Nut Lei Hibiscus Hand Painted Turquoise Flower 32 Inches. What is the most likely intention for buying them?\newline
A. because the person wants to replace the LCD screen of the laptop.\newline
B. because they both are capable of giving him a lot of benefits.\newline
C. because they both are a type of toys.\newline
D. because they could both be found on Aliexpress.com.\newline
Formulate your answer in this way.\newline
Step 1: Give a short and brief rationale by thinking step by step.\newline
Step 2: Answer A or B or C or D only without any other word.\newline
\underline{[Response from GPT-4]}\newline
Step 1: The customer is buying two different styles of Kukui Nut Lei, which are Hawaiian-style accessories. The intention is most likely related to their use as accessories or decorations.

Step 2: D\ \ \
\underline{[Correct answer: B]}\\\midrule[0.75pt]
Task 2 & 
\underline{[Prompt]}\newline
A customer buys StarTech.com 60x10mm Replacement Ball Bearing Computer Case Fan w/ TX3 Connector - 3 pin case Fan - TX3 Fan - 60mm Fan, because they both are defined as "Computer Accessories".\newline
What is the customer's most probable additional purchase?\newline
A. StarTech 8-Inch 24 Pin ATX 2.01 Power Extension Cable (ATX24POWEXT)\newline
B. Genuine Power On-Off On/Off On Off Flex Cable Ribbon FOR Apple iPad 2 iPad2\newline
C. WiFi Bluetooth Antenna Flex Cable Replacement Part for IPad 3 and Ipad 4\newline
D. Headphone Audio Jack Flex Cable Ribbon Replacment for Ipad Air (Black)\newline
Formulate your answer in this way.\newline
Step 1: Give a short and brief rationale by thinking step by step.\newline
Step 2: Answer A or B or C or D only without any other word.\newline
\underline{[Response from GPT-4]}\newline
Step 1: The customer has purchased a computer case fan, which indicates that they are likely working on a desktop computer or building one. Therefore, the most probable additional purchase would be related to desktop computer components or accessories.

Step 2: A\ \ \
\underline{[Correct!]}
\\\midrule[0.75pt]
Task 2 & 
\underline{[Prompt]}\newline
A customer buys Nakamichi NK2000 Over-the-Head Headphones | White, because he wanted to use them with his iPod.\newline
What is the customer's most probable additional purchase?\newline
A. Nakamichi Amplified Stereo Headphones NK2000 Turbulence\newline
B. Apple iPod nano 8 GB Silver (4th Generation)  (Discontinued by Manufacturer)\newline
C. Iwoo Bluetooth Headphone, Blue\newline
D. Portable CD Cassette Boombox with AM/FM Radio\newline
Formulate your answer in this way.\newline
Step 1: Give a short and brief rationale by thinking step by step.\newline
\underline{[Response from GPT-4]}
\newline
Step 2: Answer A or B or C or D only without any other word.\newline
Step 1: The customer already bought headphones for his iPod, so he is unlikely to buy another pair of headphones or a different audio device. He might need an accessory that is compatible with his iPod.\newline
Step 2: B\ \ \
\underline{[Correct answer: A]}
\\
\bottomrule[1.5pt]
\end{tabularx}
}
\caption{Example prompts and responses from GPT-4 with \textsc{CoT} prompting methods. }
\label{tab:case}
\end{table*}

\section{Error Analysis examples}
We present examples of erroneous responses by GPT-4 with CoT prompting on both tasks for each error type in Table~\ref{tab:error-task1} and Table~\ref{tab:error-task2}.
\begin{table*}[t]
\small
\begin{tabularx}{\textwidth}{>{\raggedright}p{2cm}>{\raggedright}p{3cm}>{\RaggedRight\arraybackslash}p{10cm}}
\toprule[1.5pt]
\textbf{Error Type} & \textbf{Description} & \textbf{Example}\\ \midrule[0.75pt]
\textbf{Atypical} & The selected option is not the most typical intention given the products. & 
[Question] A customer buys Invicta Men's 1515 I Force Collection 18k Gold Ion-Plated Watch with Black Cloth-Covered Band and Invicta Men's 16011 S1 Rally; 18k Gold Ion-Plated Watch with Brown Leather Strap. What is the most likely intention for buying them?\newline
A. because they both are a manner of the same brand.\newline
B. because they both are related to watches.\newline
C. because they could both be used to tell time.\newline
D. because they both are capable of providing him with the comfort and support he needs.\newline
[Response] Step 1: The customer is likely buying both watches because they are from the same brand and can be used to tell time, but they have different styles (one with a black cloth-covered band and the other with a brown leather strap).\newline
Step 2: A\newline
[Correct answer: C]\\\midrule[0.75pt]
\textbf{Overrarching Inference} & The selected options, while seemingly plausible, cannot be deduced from the products provided.& 
[Question] A customer buys KabelDirekt (20 feet) CL3 HDMI Cable (1080p 4K 3D High Speed with Ethernet ARC rated for in wall installation) - TOP Series and BlueRigger In-Wall High Speed HDMI Cable - 15 Feet - CL3 Rated - Supports 4K, Ultra HD, 3D, 1080p, Ethernet and Audio Return (Latest Standard). What is the most likely intention for buying them?\newline
A. because the person wants his home theater system to be better.\newline
B. because the person wants to record his voice and video.\newline
C. because the person wants to connect the device to the TV.\newline
D. because they both are defined as ``Network Cable'' and ``Network Adapter'' in the same category.\newline
[Response] Step 1: The customer is buying two different HDMI cables, both of which are rated for in-wall installation and support high-quality video and audio transmission. This suggests that they are looking to improve their home theater system by connecting devices to their TV.\newline
Step 2: A\newline
[Correct answer: C]
\\\midrule[0.75pt]
\textbf{Implausible} & The model selects an option that is irrelevant to the given products or implausible. Cases where the rationales in \textsc{CoT} responses are irrelevant to selected options are also observed.& 
[Question] A customer buys 14 Karat Yellow Gold Round Bead Ball Stud Earrings, 2mm and 14 Karat White Gold Round Bead Ball Stud Earrings, 3mm. What is the most likely intention for buying them?\newline
A. because they both are capable of being worn for a long time.\newline
B. because they both are a type of USB extension cable.\newline
C. because the person wants to use them together.\newline
D. because they both have a property of anti-glare/anti-fingerprint.\newline
[Response] Step 1: The customer is buying two different types of gold earrings, which indicates they are likely purchasing them for personal use or as a gift. The other options do not relate to earrings or their intended use.\newline
Step 2: C\newline
[Correct answer: A]\\\midrule[0.75pt]
\textbf{Inaccurate Product Understanding} & The selected options do not align well with the product.& 
[Question] A customer buys Sony Clip-on Stereo Headphones With Double Retractable Cord | MDR-Q68LW L Blue (Japanese Imports) and Sony Clip-on Stereo Headphones With Double Retractable Cord | MDR-Q68LW T Brown (Japanese Imports). What is the most likely intention for buying them?\newline
A. because they both are defined as ""Headset"" in the product name.\newline
B. because the person wants to listen to music or watch a movie.\newline
C. because they both are capable of being worn with a variety of outfits.\newline
D. because they both are capable of blocking out noise.\newline
[Response] Step 1: The customer is buying two different colors of the same product, which indicates they might want to match them with different outfits or have options for style.\newline
Step 2: C\newline
[Correct answer: B]\\
\bottomrule[1.5pt]
\end{tabularx}

\vspace{3pt}
\caption{Examples of erroneous responses by GPT-4 with CoT prompting methods on \taskone{}.}
\label{tab:error-task1}
\end{table*}

\begin{table*}[t]
\small
\begin{tabularx}{\textwidth}{>{\raggedright}p{2cm}>{\raggedright}p{3cm}>{\RaggedRight\arraybackslash}p{10cm}}
\toprule[1.5pt]
\textbf{Error Type} & \textbf{Description} & \textbf{Example}\\ \midrule[0.75pt]
\textbf{Inaccurate Intention Understanding} & The selected options do not align with the given intention. & 
[Question] A customer buys Nakamichi NK2000 Over-the-Head Headphones | White, because he wanted to use them with his iPod.
What is the customer's most probable additional purchase?\newline
A. Nakamichi Amplified Stereo Headphones NK2000 Turbulence\newline
B. Apple iPod nano 8 GB Silver (4th Generation)  (Discontinued by Manufacturer)\newline
C. Iwoo Bluetooth Headphone, Blue\newline
D. Portable CD Cassette Boombox with AM/FM Radio\newline
[Response] Step 1: The customer already bought headphones for his iPod, so he is unlikely to buy another pair of headphones or a different audio device. He might need an accessory that is compatible with his iPod.\newline
Step 2: B\newline
[Correct answer: A]\\\midrule[0.75pt]
\textbf{Inaccurate Product Understanding} & The reasoning in their response demonstrates inaccurate understanding of the purchased products or those in the options. Or, when the intention is not typical enough to filter out distractors, they fail to rely more on the purchased product to select the best option.& 
[Question] 
A customer buys Lapis ID Necklace, because they could both be used as a necklace.
What is the customer's most probable additional purchase?\newline
A. The Lord Fashion Jewelry Arwen Evenstar Twilight Star Pendant Necklace\newline
B. Sterling Silver Antiqued Tennessee State Charm\newline
C. Juvel Jewelry Women Fashion 18K Gold Plated Stud Earrings Elegant Cubic Zircon Ring Style With Diamond For Party\newline
D. Ms. Sophisticated ID Necklace\newline
Formulate your answer in this way.\newline
[Response] Step 1: The customer is most likely to purchase another item that can also be used as a necklace and complements their initial purchase.\newline
Step 2: A\newline
[Correct answer: D]
\\
\bottomrule[1.5pt]
\end{tabularx}

\vspace{3pt}
\caption{Examples of erroneous responses by GPT-4 with CoT prompting methods on \tasktwo{}.}
\label{tab:error-task2}
\end{table*}

\end{document}